# ChatGPT and general-purpose AI count fruits in pictures surprisingly well

## Authors

Konlavach Mengsuwan[1, 2], Juan Camilo Rivera Palacio[1,2,3], Masahiro Ryo[1,2,*]

## Affiliations

[1] Research Platform "Data Analysis & Simulation", Leibniz Centre for Agricultural Landscape Research (ZALF), Müncheberg, Germany

[2] Environment and Natural Sciences, Brandenburg University of Technology Cottbus‐Senftenberg, Cottbus, Germany

[3] Alliance of Bioversity International and CIAT, Rome, Italy.

* corresponding author: Masahiro Ryo (Masahiro.Ryo@zalf.de)

## Abstract

Object counting is a popular task in deep learning applications in various domains, including agriculture. A conventional deep learning approach requires a large amount of training data, often a logistic problem in a real-world application. To address this issue, we examined how well ChatGPT (GPT4V) and a general-purpose AI (foundation model for object counting, T-Rex) can count the number of fruit bodies (coffee cherries) in 100 images. The foundation model with few-shot learning outperformed the trained YOLOv8 model ($R^2$ = 0.923 and 0.900, respectively). ChatGPT also showed some interesting potential, especially when few-shot learning with human feedback was applied ($R^2$ = 0.360 and 0.460, respectively). Moreover, we examined the time required for implementation as a practical question. Obtaining the results with the foundation model and ChatGPT were much shorter than the YOLOv8 model (0.83 hrs, 1.75 hrs, and 161 hrs). We interpret these results as two surprises for deep learning users in applied domains: a foundation model with few-shot domain-specific learning can drastically save time and effort compared to the conventional approach, and ChatGPT can reveal a relatively good performance. Both approaches do not need coding skills, which can foster AI education and dissemination.

# Introduction

ChatGPT caused a technological breakthrough in the artificial intelligence (AI) domain and beyond (van Dis et al., 2023). ChatGPT is a large language model (LLM) to hold a humanlike conversation with the end user, released by OpenAI in November 2022. The ChatGPT has notable features for various tasks at an unprecedented level, which can be operated with only human language prompting (Zhang & Shao, 2024). Beyond ChatGPT, the potential of LLM applications has been actively explored for specific contexts in science, technology, and society, including plant science (Yang et al., 2024) and agriculture (Tzachor et al., 2023). Another trend in the AI domain is to develop general-purpose AI, also known as foundation models. Foundation models are "the models which are trained on broad data that can be adapted to a wide range of downstream tasks" (Bommasani et al., 2022). Overall, it is increasingly common to develop a single large AI model that can be used for various tasks, and several of them do not require any programming skills.

In agricultural science, deep learning techniques have been implemented intensively for computer vision and image recognition across a variety of tasks such as plant health diagnosis and phenotyping, field status monitoring, and crop yield prediction (Kamilaris & Prenafeta-Boldú, 2018; Ryo et al., 2023). However, the potential of the recent AI advancement with LLM and foundation models has not been explored for computer vision tasks in agricultural science. To date, an uncountable number of models have been developed, and they are mostly specialized for a very narrow task based on a given training data (Bouguettaya et al., 2022; Farjon et al., 2023; Saleem et al., 2021). Developing every single model requires a good programmer with a considerable amount of data collection and annotation. These requirements are a key obstacle for upscaling a promising AI model demonstrated in a particular context to a broad application under various contexts.

We show that the recent development of LLMs and foundation models lowers the hurdle drastically, especially for developing a context-specific image analysis without requiring any model training effort and programming skills. For this demonstration, we tested the potential of GPT-4V (large language-vision model by OpenAI) and T-Rex (foundation model for counting objects in an image (Jiang et al., 2023)) for counting coffee cherry fruit bodies in images taken by farmers in their farms. GPT-4V just needs an image and a query like "count the number of fruit bodies in the image" (technically, zero-shot learning). We also tested if the model performance improves by providing feedback. while T-Rex requires drawing a few bounding boxes on an image to instruct what to count (few-shot learning). We compared these modeling approaches to a conventional deep learning approach in which we trained the current best object detection algorithm, YOLO v8, in the previous study (Palacio et al., 2024) (**Fig. 1**). More specifically, we tested their performances based on 100 images of coffee tree branches with cherries taken with local farmers' mobile phones in Colombia, where the number of cherries was counted manually by them (ranging from 1 to 80 cherries).

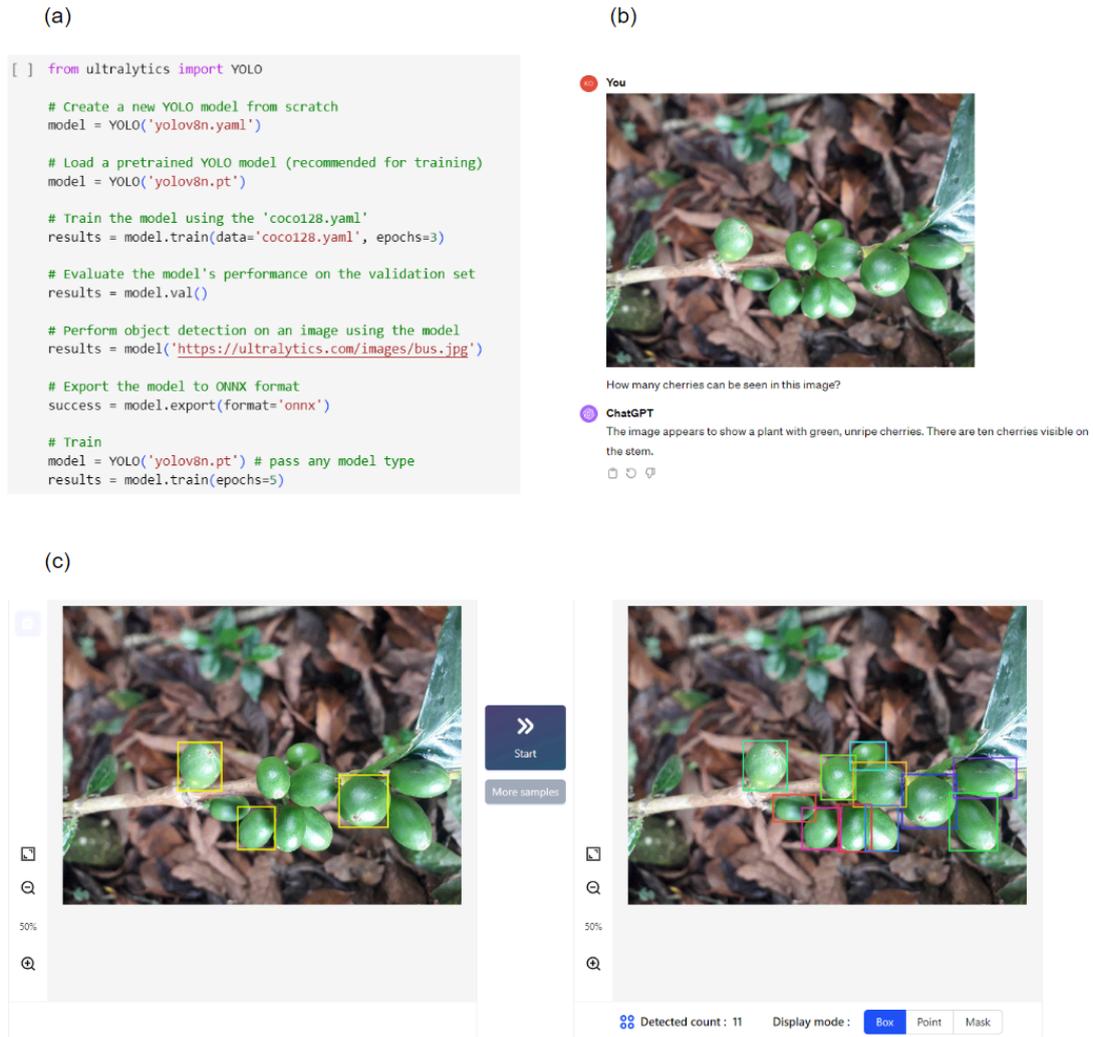

**Figure 1**. **Conventional deep learning, ChatGPT, and foundation model count the number of cherries**. (a) an example of standard deep learning prediction model implementation in Python, which requires coding and data annotation with a large number of images. (b) ChatGPT (GPT-4V) needs only a query. In this example, we asked, "How many cherries can be seen in this image?", and the model answered "10" along with describing what the image is about. (c) T-Rex is a foundation model for counting objects, which requires the user to draw a few bounding boxes for instruction. In this example, after instructing what to count (left), the model identified "11" cherries in the image (right).

## Results

The trained task-specific model, YOLOv8, exhibited robust prediction capabilities with an $R^2$ score of 0.900 (as a benchmark; **Fig. 2a, 3a**). In comparison, the foundation model specific for object counting, T-Rex, surpassed this performance ($R^2$ = 0.923; Fig. 2b, 3b), demonstrating its superior predictive efficacy. The GPT-4V recorded $R^2$ scores of 0.360 by just requesting to count the number of cherries (**Fig. 2c, 3c**). Yet, the performance was increased substantially by giving feedback by the user indicating under- or overestimation ($R^2$ = 0.460; **Fig. 2d, 3d**). Both GPT-4V approaches underestimated cherries, and they did not suggest numbers higher than 45 for any images, which was not observed with YOLOv8 and T-Rex.

The T-Rex model performed as the most time-efficient, completing the 100 images dataset analysis in approximately 0.83 hours (**Fig. 3**). The GPT-4V required a slightly extended period of 1.75 hours (zero-shot) and 3.25 hours (few-shot) because of the restricted use (40 messages per 3-hour limit as of March 2024). The YOLOv8 required 161 hours, including the annotation of 500 images with 30000 cherries for preparing the training dataset, model training, and the administrative process for hiring personnel to do annotation.

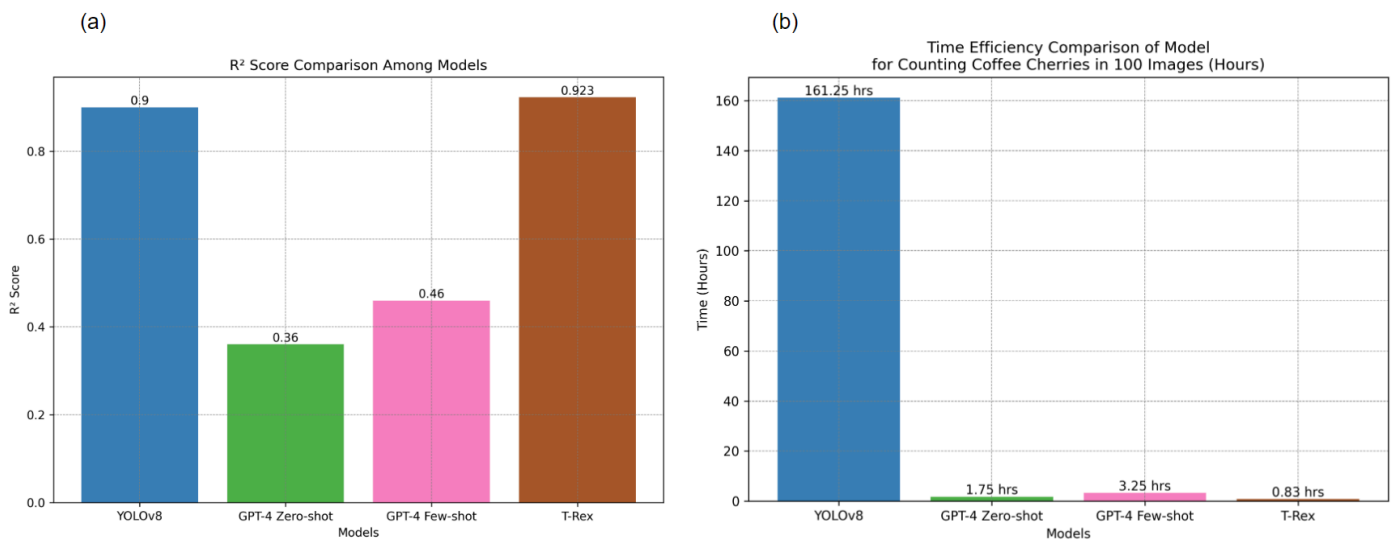

**Figure 2. Performance comparison among conventional deep learning, ChatGPT, and Foundation model.** Models were compared in terms of (a) performance (R-squared between observed and predicted number of cherries seen in the image) and (b) the total time required for completing counting the number of cherries since having the 100 images of coffee tree branches. YOLOv8 includes the manual image annotation task for YOLOv8 under the assumption that the counting task needs to be done from scratch while having a large number of images. T-Rex, the foundation model without training, surpassed the performance of the YOLOv8 model, the state-of-the-art object detection algorithm with training. GPT-4 has also demonstrated some ability to count the number of cherries.

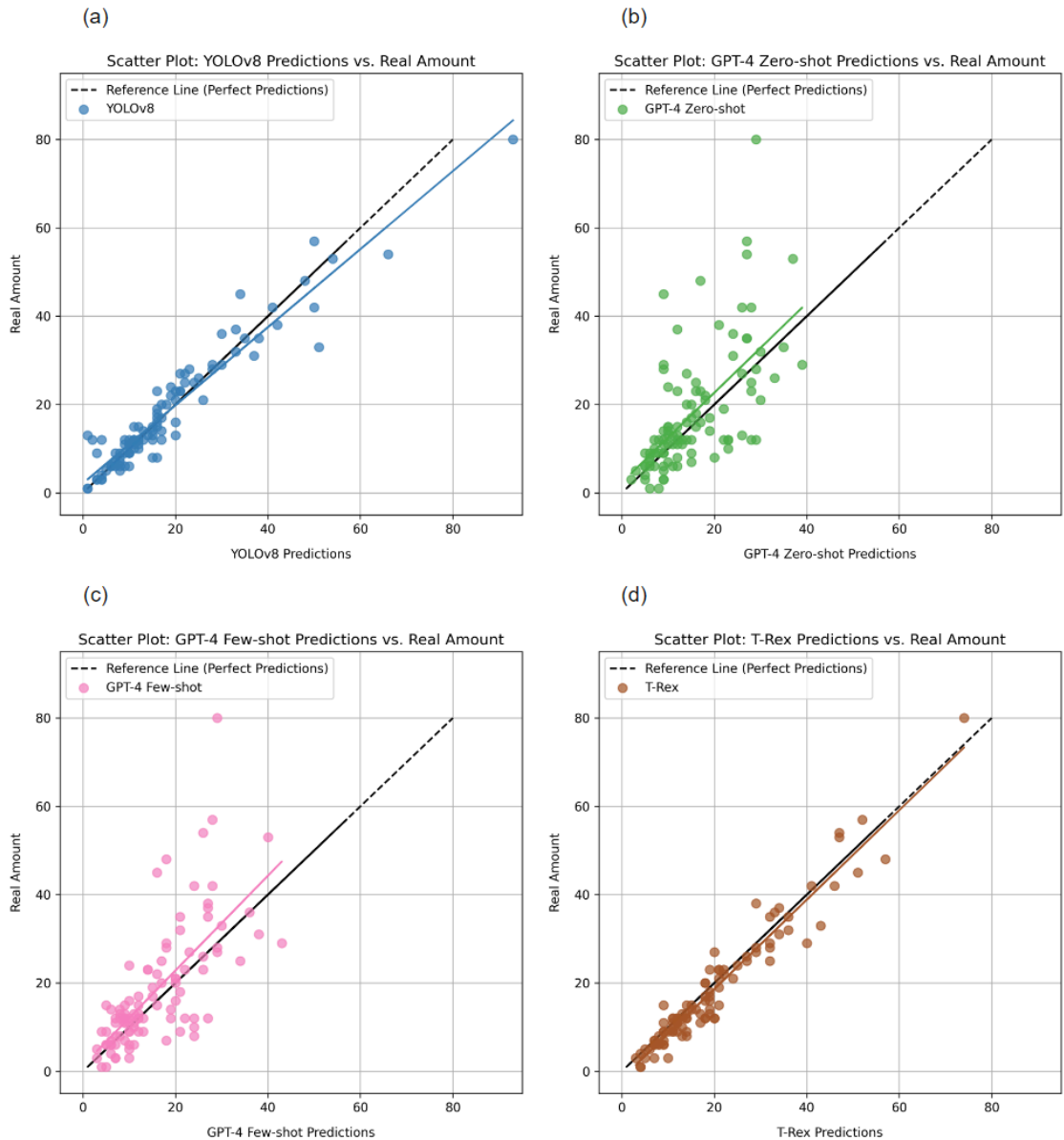

**Figure 3** Scatter plots comparing predictions and observations. (a) YOLOv8, (b) GPT-4 Zero-shot learning, (c) GPT-4 with providing user feedback, and (d) T-Rex, the foundation model for object counting.

## Discussion

GPT-4V and T-Rex have great potential, even though there are still some caveats and limitations in their applicability and efficiency. The T-Rex model, the foundation model for object counting without training, surprisingly outperformed the state-of-the-art YOLOv8 model with training. The finding is striking because employing a foundation model for object counting has not been mentioned in the latest review article about deep learning-based object counting methods in agricultural science (Farjon et al., 2023). We imagine that the application of foundation models beyond object counting for various tasks could be a major approach in deep learning applications, including crop yield prediction (Coviello et al., 2020;

Gutiérrez et al., 2019; Häni et al., 2020; Maheswari et al., 2021; Mekhalfi et al., 2020; Saddik et al., 2023), fruit health detection (Khattak et al., 2021; Momeny et al., 2022; Uğuz et al., 2023), immature/mature detection (Lu et al., 2022; Tenorio & Caarls, 2021), and chemical usage planning (Farjon et al., 2020), especially where traditional data labeling and processing methods are resource-intensive or difficult to implement. Also, no programming skill is required.

ChatGPT also revealed interesting potential, although the performance was clearly worse than the T-Rex and YOLOv8 models. To our knowledge, this is the first attempt to estimate the ability of GPT-4V for an object-counting task in a well-designed experiment. It is interesting to confirm that just offering prompt feedback if it was over-/under-estimated can improve the model performance. Nevertheless, unlike the foundation model, securing reproducibility is a challenge when employing LLMs like GPT-4V for academic research (Franc et al., 2024; May et al., 2024). The inability to set a random seed or control certain parameters during the learning process, which is the scientific standard, could pose significant challenges in scenarios where consistent results are essential. This limitation underscores the need for ongoing research and development to address these methodological constraints. Yet, the speed of development is astonishingly fast, and therefore, this issue could be solved in the future.

In terms of scalability, YOLOv8 stands out as a winner due to its speed and lower energy demands once trained, especially as the model size is small (Palacio et al., 2024). This makes it a suitable choice for scenarios where rapid processing is needed, a large number of photos need to be processed, and low-energy consumption is critical (e.g. implementing the model in an edge device in a low-income country). However, when considering user-friendliness, the T-Rex model stands out. Its simple user interface and the ability to select specific object types make it an attractive option, especially in user-centric applications where simplicity is a key. Moreover, it is noted that there has been substantial progress in reducing the size of a large neural network model (e.g. BitNet; Ma et al., 2024).

The domain-specific challenges in agricultural applications, as identified by (Farjon et al., 2023), include target object variability, variations in illumination, object occlusions, double counting, and distinguishing between background and foreground. Traditional deep learning models, while robust in structured environments, often falter when confronted with these agricultural-specific challenges due to their reliance on consistent, high-quality datasets. In contrast, foundation models like T-Rex exhibit a degree of adaptability to object variability and occlusions, benefiting from a broader learning base. GPT-4V, with its advanced understanding and contextual capabilities, could offer innovative solutions to these problems, particularly in interpreting complex scenarios involving illumination and occlusions.

In conclusion, foundation models hold considerable promise for object-counting tasks, and multimodal LLMs have been showing some interesting potential, but their selection should be guided by specific requirements such as precision, scalability, energy cost, and user-friendliness. Task-specific models like YOLOv8 and foundation models, including T-Rex, offer advantages in terms of speed and precision, while general LLM models like GPT-4 provide ease of use and flexibility. The future of these technologies lies in their continued development and adaptation, making them more versatile and efficient across various applications.

## Materials and Methods

The mobile pictures used in this study were taken as a subset from the larger data collection in the previous study (Palacio et al., 2024). Here, we shortly summarize the key procedures for data collection and model development. The images were collected from 300 local partners during the cherry growing season from March to November 2022 in the Cauca and Quindio departments. The species studied was Arabica coffee (*Coffea arabica L.*), including varieties such as *variedad supremos*, *castillo*, and '*variedad Colombia*. The images were taken with basic smartphones owned by the partners in natural lighting conditions in real coffee growing environments. Three pictures were taken per tree, positioned at the top, middle, and bottom, taking into account the following considerations: capturing pictures during daylight hours, focusing on as many cherries as possible, avoiding immediate movement of the camera after capturing the picture, and positioning the camera so that sunlight does not shine directly on it. For this study, we selected the subset (100 images), representing a diverse real scenario in the coffee crop, including ensuring no blurriness or distortion, capturing different positions of the camera, featuring different varieties, using the screen to reveal hidden cherries, and capturing objects resembling coffee cherries.

The YOLO v8 (Terven & Cordova-Esparza, 2023) model was trained using an annotated dataset comprising 436 images, which collectively contained 35,694 labeled cherries in the previous study (see Palacio et al., 2024 for the details). Among these cherries, 35,247 (98.7%) were labeled as green, 342 (0.9%) as red, and 105 (0.2%) as black. All models were developed on a Windows 10 platform using Python 3.7.0 routines running on an Nvidia GeForce RTX 3080 GPU clocked at 1440 MHz. Note that the 100 images used in this study were not used for training the model.

We used ChatGPT-4 (*ChatGPT-4*, 2023) on December 7, 2023, for the analysis. As zero-shot learning and few-shot learning, groups of three images were uploaded to the ChatGPT-4 interface, each followed by a prompt *"Could you count the coffee cherry which is visually seen in these 3 images carefully please? Please provide the result as a table."* for the model to count visible cherries, the results of which were carefully collected and documented in a spreadsheet. This process was repeated across the entire set without any iterative feedback to ensure the model's initial unbiased performance was accurately recorded. Moreover, to give feedback, in the few-show learning approach, a custom GPT model was created and provided the following instructions: *"The GPT, is guided with a dataset of 10 images and is primed for real-case scenarios. It is committed to delivering the most accurate cherry counts possible from the images provided. The GPT should carefully consider about individual cherries, accounting for* *clustering, varying ripeness levels, lighting, and background contrasts. The GPT will consistently apply its best analysis on every image, using the knowledge gained from the examples to ensure the counts are as precise as they can be.*" We kept the *"Web Browing"* and *"*DALL·E Image Generation" capabilities on, and we turned the "Code Interpreter" capability off. We then uploaded a separate dataset of 10 images with known real count one by one to the GPT Builder and provided the prompt as follows: *"Could you count the coffee cherry which is visually seen in an uploaded image carefully please?"* and run the GPT Builder. After the GPT Builder provided the count, if the count from GPT was significantly incorrect, we then provided the real amount and requested the GPT to learn if there were any difficulties in counting the specific image, as the following prompt: "*The real amount is x. Please investigate and learn what made you count the cherries inaccurately. Please learn and overcome those difficulties.*", otherwise, a prompt was provided with real count, commend the GPT's ability, and to mention that the counting approach was on the right track. After guiding through these ten iterative feedback cycles, this Custom GPT saved and created a new GPT-4 chat with this Custom GPT. The uploading images, prompting, and documenting process were similar to the zero-shot learning.

The T-Rex model's (Jiang et al., 2023) object counting proficiency was also used. Using the online demo available on the Deep Data Space Playground, each image was uploaded individually. Three boundary boxes were manually drawn around visible coffee cherries in each image to guide the T-Rex model in identifying and counting the cherries. The detected counts were then recorded for each image, and this procedure was systematically applied to the entire dataset of 100 images.

These four model approaches were evaluated based on their accuracy and total time required in counting coffee cherries in the 100 images. Accuracy was quantified using the R-squared value, calculated with the 'R2_score' function from the 'sklearn.metrics' library version 1.2.2 (Pedregosa et al., 2011) in Python programming language version 3.10.12 (Van Rossum & Drake, 2009). Time efficiency was evaluated, including time for uploading images, writing prompts, waiting to receive results, providing guidance with additional images (for the few-shot learning only), annotation (for the YOLOv8 model only), and training (YOLOv8 only). The time required for obtaining predictions for the 100 images showed significant contrasts in processing efficiency. The ChatGPT-4 Zero-shot learning approach required approximately 105 minutes, including uploading images, writing prompts, GPT-4 analyzing, and documenting results, while the Few-shot approach required about 195 minutes, including additional time for providing guidance with ten additional images in Custom GPT. The T-Rex model needed about 50 minutes to process all images, allocating approximately 30 seconds per image for uploading, drawing 3 bounding boxes, analyzing, and documenting results. For the YOLOv8 model, the annotation process required 160 hours for 436 images containing 35,694 labeled cherries, the run-time for the model training took 75 minutes, and the average time predictions per picture was 30 milliseconds.

# Acknowledgment

This work was supported by the Federal Ministry of Education and Research (BMBF – Bundesministerium für Bildung und Forschung) project "Multi-modal data integration, domain-specific methods, and AI to strengthen data literacy in agricultural research" (16DKWN089), WIR! - Land - Innovation - Lausitz (LIL) project "LandscapeInnovations in Lausitz (Lusatia) for Climate-adapted Bioeconomy and nature-based Bioeconomy-Tourism" (03WIR3017A), the Brandenburg University of Technology Cottbus-Senftenberg (BTU) with the Graduate Research School cluster project "Integrated analysis of Multifunctional Fruit production landscape to promote ecosystem services and sustainable land-use under climate change" (BTUGRS2018_19), and by Deutsche Gesellschaft für Internationale Zusammenarbeit (GIZ) with the Croppie project (grant number 81275837).

https://doi.org/10.1007/s11119-019-09679-1

Franc, J. M., Cheng, L., Hart, A., Hata, R., & Hertelendy, A. (2024). Repeatability, reproducibility, and diagnostic accuracy of a commercial large language model (ChatGPT) to perform emergency department triage using the Canadian triage and acuity scale. *Canadian Journal of Emergency Medicine*, *26*(1), 40–46. https://doi.org/10.1007/s43678-023-00616-w

Gutiérrez, S., Wendel, A., & Underwood, J. (2019). Ground based hyperspectral imaging for extensive mango yield estimation. *Computers and Electronics in Agriculture*, *157*, 126–135. https://doi.org/10.1016/j.compag.2018.12.041

Häni, N., Roy, P., & Isler, V. (2020). A comparative study of fruit detection and counting methods for yield mapping in apple orchards. *Journal of Field Robotics*, *37*(2), 263–282. https://doi.org/10.1002/rob.21902

Jiang, Q., Li, F., Ren, T., Liu, S., Zeng, Z., Yu, K., & Zhang, L. (2023). *T-Rex: Counting by Visual Prompting*. https://doi.org/10.48550/ARXIV.2311.13596

Kamilaris, A., & Prenafeta-Boldú, F. X. (2018). Deep learning in agriculture: A survey. *Computers and Electronics in Agriculture*, *147*, 70–90. https://doi.org/10.1016/j.compag.2018.02.016

Khattak, A., Asghar, M. U., Batool, U., Asghar, M. Z., Ullah, H., Al-Rakhami, M., & Gumaei, A. (2021). Automatic Detection of Citrus Fruit and Leaves Diseases Using Deep Neural Network Model. *IEEE Access*, *9*, 112942–112954. https://doi.org/10.1109/ACCESS.2021.3096895

Lu, S., Chen, W., Zhang, X., & Karkee, M. (2022). Canopy-attention-YOLOv4-based immature/mature apple fruit detection on dense-foliage tree architectures for early crop load estimation. *Computers and Electronics in Agriculture*, *193*, 106696. https://doi.org/10.1016/j.compag.2022.106696

Ma, S., Wang, H., Ma, L., Wang, L., Wang, W., Huang, S., Dong, L., Wang, R., Xue, J., & Wei, F. (2024). *The Era of 1-bit LLMs: All Large Language Models are in 1.58 Bits* (arXiv:2402.17764). arXiv. http://arxiv.org/abs/2402.17764

Maheswari, P., Raja, P., Apolo-Apolo, O. E., & Pérez-Ruiz, M. (2021). Intelligent Fruit Yield Estimation for Orchards Using Deep Learning Based Semantic Segmentation Techniques—A Review. *Frontiers in Plant Science*, *12*, 684328. https://doi.org/10.3389/fpls.2021.684328

May, M., Körner-Riffard, K., & Kollitsch, L. (2024). Can ChatGPT Realistically and Reproducibly Assess the
10